\documentclass[runningheads]{llncs}
\usepackage[T1]{fontenc}
\usepackage{graphicx}
\usepackage{booktabs}
\usepackage{float}
\usepackage{placeins}
\usepackage[export]{adjustbox}
\begin{document}
\title{Ordered Network Analysis of Epistemic Emotions during Collaborative Problem Solving}
\titlerunning{Ordered Network Analysis of Epistemic Emotions during CPS}
% \author{Anonymous Authors}
\author{Sifatul Anindho\inst{1}\orcidID{0009-0009-4641-4798} \and
Videep Venkatesha\inst{1}\orcidID{0009-0000-4635-3010} \and
Jaclyn Ocumpaugh\inst{2}\orcidID{0000-0002-9667-8523} \and
Nathaniel Blanchard\inst{1}\orcidID{0000-0002-2653-0873}}
\authorrunning{S. Anindho et al.}
\institute{Colorado State University, Fort Collins CO 80523, USA\\
\email{\{sifatul.anindho, videep.venkatesha, nathaniel.blanchard\}@colostate.edu}\\
\and
University of Houston, Houston TX 77104, USA\\
\email{jlocumpaugh@gmail.com}
}
\maketitle              % typeset the header of the contribution
\begin{abstract}
Investigating how affective states such as confusion and frustration persist and transition during co-situated collaborative problem solving (CPS) is important for understanding the dynamics of these epistemic emotions. However, the accurate identification of affective states remain challenging as there is no gold-standard truth in this space. Here, we analyze affective states collected through retrospective cued-recall during an in-person CPS task. Using ordered network analysis (ONA), we examine (1) the overall ordered structure of affective states and how this structure differs across self-caught and probe-caught reporting methods, and (2) what aspects of this ordered structure are emphasized differently in slower and faster groups. We find that ONA reveals differences in persistence and transition patterns that are not apparent from descriptive summaries alone. In particular, we observe a stable epistemic core linking curiosity, optimism, and confusion, with different reporting methods emphasizing different connections among states. An analysis between faster and slower groups show that roles of confusion and disengagement also shift significantly during collaboration, particularly in their relationship to conflict. We interpret our findings in the context of collaboration and discuss their implications in developing AI systems that support CPS.

\end{abstract}
\section{Introduction}
Collaborative problem solving (CPS) is an important part of modern education, requiring learners to coordinate ideas, communicate effectively, and regulate cognitive and affective processes while working toward shared goals \cite{hesse_cps}. Research has shown that affective states such as confusion and frustration play an important role in learning \cite{dmello_dynamics}. Consequently, there is interest within the AIED community in developing systems that can perceive and adapt to learners’ affective states \cite{ismail_emotionally_adaptive_2023,yu_affect_prediction_2024}.

Previous work has focused on individual learning or computer-mediated collaboration, often using self-reports \cite{pekrun_emotions_measurement_2011}, log data \cite{baker_bromp_2020}, physiological data \cite{hussain_autotutor_physiology_2011}, discourse features \cite{dmello_language_emotions_2012}, think-aloud protocols \cite{zhang_llm_think_aloud_srl_2024} or observable behaviors \cite{bosch_video_affect_2016} as signals for these states. Although valuable, think-aloud protocols and physiological detectors risk disrupting naturalistic in-person CPS, while log-based measures are often insufficient to capture affective processes in short co-situated collaborative tasks. Traditional self-reporting methods are prone to recall bias and lack temporal resolution \cite{akbulut_self_reports_bias_2025}, and discourse/non-verbal observable signals are difficult to map onto the internal experiences of learners during collaboration due to social masking \cite{underwood_social_masking_1997}. Consequently, affective states in collaboration are not directly observable, and no single measurement approach can be treated as a definitive ground truth.

To avoid task interference, researchers often use cued-recall procedures \cite{russell_retrospective_cued_recall_2009}, where participants reconstruct the memory of their experiences by reviewing recordings of their activity. While this has been shown as a robust method for eliciting affective states in some settings \cite{bentley_rcr_gaming_2005}, its application in short-duration CPS faces scalability and granularity challenges. User studies are often lengthy and the precise temporal fidelity of these reports is often difficult to determine, as the exact onset and offset of a state may not perfectly align with what people remember. Because of the limitations in size and perceived lack of granularity, researchers often default to descriptive summaries, which implicitly treat affective states as independent \cite{anindho_internal_states_2025}. This approach overlooks the sequential transitions and co-occurrences inherent in the data. Even with limited temporal granularity and scale, these reports may contain quantifiable information about the order and flow of epistemic states; there is a clear need to move beyond summary statistics to explore these transitions in temporally coarse, small-scale educational datasets.
 
This paper investigates the ordered structures of affective states using a dataset collected from small groups engaged in a short situated CPS task using retrospective cued-recall (RCR). During the recall phase, participants reported affective states using RCR, either voluntarily at any point (self-caught) or being probed after a fixed period of inactivity (probe-caught). We analyze these self-reports using ordered network analysis (ONA) \cite{tan_ona_2023} to examine ordered associations among affective states within short reporting windows. 
ONA allows us to model local temporal structure by capturing directional relationships between states reported earlier and later in time.

We address two research questions: (1) how are affective states structured and ordered during situated CPS as captured through self-caught and probe-caught RCR, (2) how does the ordered structure of affective states differ between faster and slower groups?
Our findings offer insight into the ordered organization of affective experience in situated CPS and how instrumentation choices shape these structures, as well as how affective organization varies with task duration. 
\section{Related Work}
\vspace{-2pt}
Research has shown that affective states are integral to learning. Psychological theories such as control–value theory describe how emotions arise from learners’ perceptions of control and value and influence learning-related behavior \cite{pekrun_control_value}. Related work on epistemic and learning-centered affect has shown that states such as confusion, frustration, and surprise commonly emerge during sense-making and problem solving, and that their role depends on how they collectively unfold over time rather than their independent presence \cite{dmello_half_life_2011}. However, as Karumbaiah et al. \cite{karumbaiah_self_transition_2019} and others have shown, aggregating affective experiences across full activities has not produced strong empirical evidence for these theoretical models, leading some to wonder whether other temporal dynamics and constructs may be important \cite{ocumpaugh_sdvet_2025}. 
Moreover, it is not clear that the emotions typically studied in individualized learning activities fully capture the range of affective experiences that may occur in collaborative activities. For example, in individual activities, research tends to focus on a handful of fully-internal emotions that have been shown to inhibit learning in some cases \cite{baker_better_frustrated_2010}, but which sometimes have contradictory effects depending on their duration \cite{lui_sequences_frustration_2013}. In such settings, dense log data or repeated measurements of observable and physiological signals is practical in tracking affective dynamics. Researchers have modeled transitions among confusion, frustration, and engagement during complex computer-mediated learning tasks, showing that short-lived affective episodes can support or hinder learning depending on context \cite{dmello_dynamics}. However, these approaches are less suited for short, situated, in-person collaboration, where interaction unfolds quickly and opportunities for dense measurement are limited. While socio-emotional interactions and shared physiological arousal events have been examined in collaborative problem solving (CPS) \cite{nguyen_ssr_physiological_2023,dindar_physiological_arousal_cps_2022}, few
studies capture affective states directly, and those often rely on external coding or sensors that risk distorting
authentic interaction \cite{ma_confusion_detection_cl_2024}. 

Recent work has used retrospective cued-recall (RCR) to elicit learners’ internal states during co-situated CPS without interrupting the flow of interaction, enabling richer access to participants’ subjective experiences than in-situ probes or post-hoc surveys \cite{anindho_internal_states_2025}. However, because such datasets are typically small and temporally coarse, analyzing the structure and evolution of affective states remains challenging. Traditional analyses of similar datasets often emphasize frequencies or aggregate transition metrics (e.g., the L-statistic \cite{dmello_dynamics}), which can obscure how such states co-occur locally and are organized over short temporal spans. Epistemic network analysis (ENA; Shaffer et al., 2016) \cite{shaffer2016quantitative} was introduced to address similar challenges by modeling learning as a network of co-occurring elements, and has since been applied to study affective dynamics \cite{karumbaiah_ena_affect}, offering a reliable way to compare patterns of connections across conditions or groups. However, standard ENA collapses temporal order within analytic windows, making it less suited for capturing directional or sequential dynamics among internal states. To address this limitation, ordered network analysis (ONA) can be used to incorporate temporal ordering into ENA network construction, allowing edges to represent ordered transitions between codes rather than unordered co-occurrence \cite{tan_ona_2023}. This makes ONA particularly well suited for modeling affective dynamics, where the sequence in which states arise may be as informative as their co-presence. 
\section{Methods}

\subsection{Data Collection}
\vspace{-2.9pt}
This study examines 27 participants, all of whom were over 18 years old and fluent in English. Recruitment was conducted within the authors’ department, and participants were compensated USD 15 with approval from the Institutional Review Board. Table \ref{tab:dem_data} summarizes the aggregated demographic information reported by participants, and Figure 1 shows three team members collaborating during an experimental task. 
\begin{table}[ht]
\centering
\caption{Demographic information of participants in our study}\label{tab:dem_data}
\vspace{3mm}
\begin{tabular}{lccc}
\toprule
\textbf{Gender} & Male (18) & Female (7)  & Non-binary (2) \\
\midrule
\textbf{Native Language} & English (17) & Other (10) & \\
\midrule
\textbf{Age} & 18--24 (14) & 25--34 (11) & 35+ (2) \\
\midrule
\textbf{Ethnicity} & Asian (13) & White (12) & Hispanic (2)\\
\bottomrule
\end{tabular}
\end{table}
\begin{figure}
\includegraphics[width=0.6\textwidth,  center]{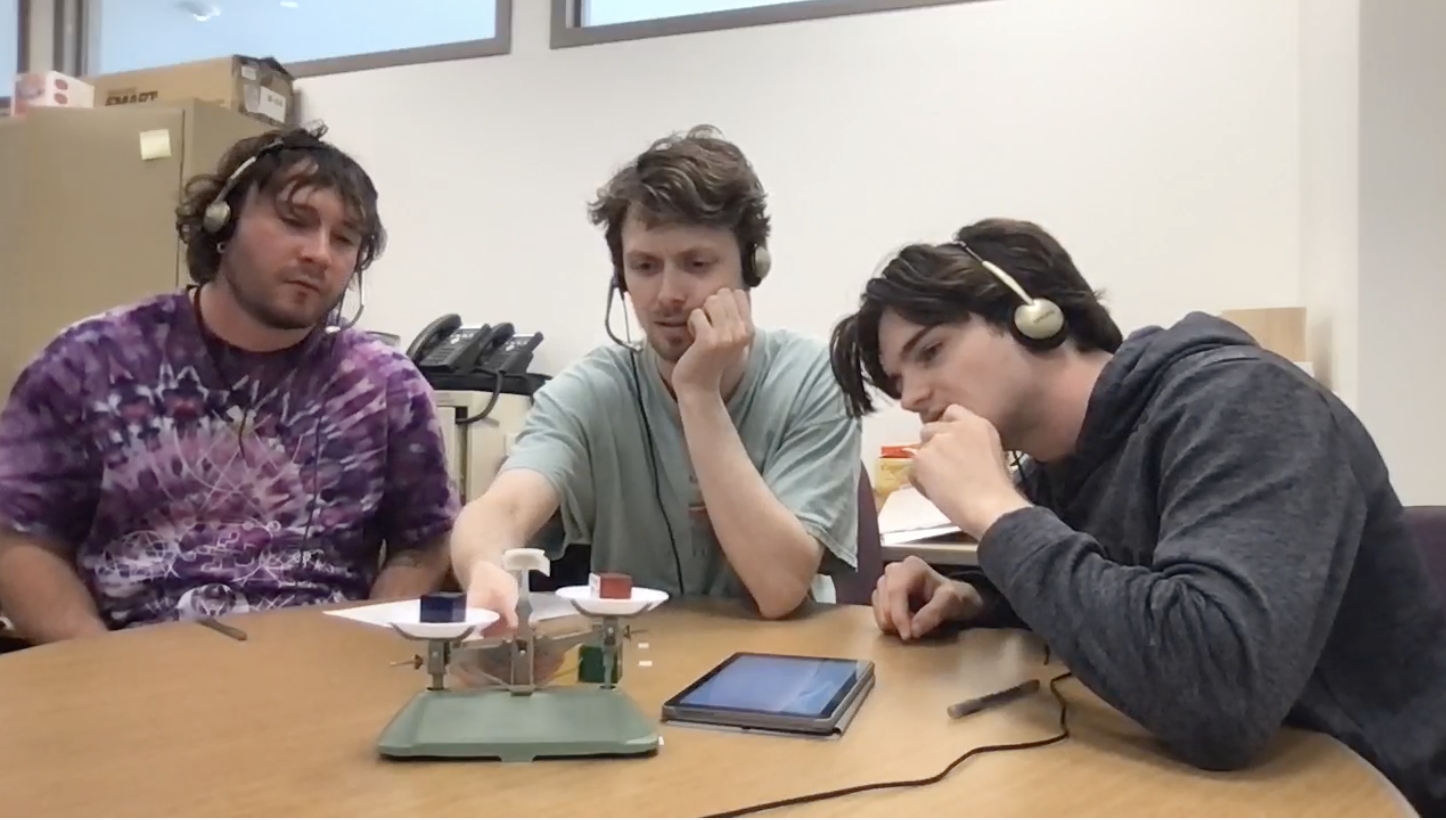}
\caption{Our experimental setup during a collaborative problem solving session} 
\label{experimental-setup}
\end{figure}
Participants were organized into 9 groups of three, and then video recorded as they collaborated to complete the first phase of the Weights Task, a collaborative problem solving (CPS) activity adapted from Khebour et al \cite{khebour_weights_task_2024}. In this activity, groups use a balance scale to infer the relative weights of five blocks, one of which has a known reference weight. They are required to reach consensus and submit a single group response via survey. 
Afterwards, participants individually reviewed the recording of their collaborative session using an retrospective cued-recall (RCR) procedure, where they reported their own affective states as they watched using an interactive survey tool. Reports could be self-initiated by the participant at any point during playback. However, after a fixed interval of inactivity (60s), participants would receive a probe to provide a report.  Participants could select one or more affective states from a set of seven (Confused, Curious, Frustrated, Disengaged, Optimistic, Surprised, and Conflicted) adopted from previous research \cite{anindho_internal_states_2025}. Both self-initiated and probed reports were timestamped, facilitating the analysis of affect dynamics reported in this paper. All groups arrived at the correct final solution, hence we do not use task accuracy as part of our analysis.
\subsection{Data Preprocessing}
Affect labels are standardized and consolidated into a single dataset, with each reported affective state represented as a separate entry. Reports are classified as self-caught or probe-caught based on whether they were initiated by the participant or triggered automatically after a fixed interval. To support comparative analysis of reporting mechanisms, we additionally construct two parallel datasets containing only self-caught and only probe-caught reports, respectively. To examine differences associated with task duration, we use a median split to divide groups into shorter- and longer-duration conditions. Each reporting event is associated with its corresponding video timestamp, affective label, group and participant identifiers, reporting mechanism, and task duration category.
\subsection{Data Analysis}
We first examine the frequency of each affect across groups and reporting conditions, allowing us to better ground our primary analysis. We then apply ordered network analysis (ONA) using the web-based epistemic network analysis (ENA) tool webena to examine the ordered relational structure of the seven affective states in our study. For this analyses, we use both standard ONA models as well as ONA difference models \cite{shaffer2016quantitative}, which allow us to hold the network analysis steady while comparing two datasets (e.g., comparing the networks of groups who performed the task slowly to the networks of the groups who performed the task quickly). We treat each reported affective state as a line of data, using groups as units of analysis, and each participant's sequence of responses as conversations. We use a moving window of 2 events to capture short-range and local patterns of persistence and transitions in affect, since such states unfold over brief stretches of interaction rather than instantaneously \cite{dmello_dynamics}. To assess the robustness of this choice, we conducted a sensitivity analysis using window sizes of 2, 3, and 4 events. The resulting network structures were consistent across these settings, with no substantial changes in node positioning or dominant connection strengths, suggesting that our findings are not sensitive to the specific window size used.

The ONA model first constructs directed (asymmetric) connections by modeling the temporal order of connections between codes that appear within the moving window of survey responses in a conversation. When multiple affective states were selected within the same survey response (i.e., at the same timestamp), these co-occurring labels were treated as ordered according to a fixed label sequence. This operationalization introduces an arbitrary ordering among simultaneous states; however, because the majority of ordered connections arise across successive responses rather than within a single response, this choice is unlikely to substantially affect the overall network structure. These co-occurrences are then aggregated at the unit level and normalized to account for the differences in reporting frequency across groups. The resulting networks are mean-centered and subjected to singular value decomposition, which produces orthogonal dimensions that maximize the variance explained by each dimension.  (See \cite{bowman_ena_math_2021} and \cite{shaffer2016quantitative} for a more detailed explanation of the mathematics.)

ONA models are visualized using directed network graphs where nodes correspond to the codes (affective states), and edges reflect the relative frequency of co-occurrence (or connection strength) and direction between two codes. Node size shows the strength of self-transitions for each state, with larger nodes indicating states that are more likely to persist across successive reports. The resulting diagram shows (1) a plotted point, which represents the location of that unit’s network in the low-dimensional projected space, and (2) a directed weighted network graph normalized on a scale from 0 to 1. The fixed positions of the nodes are optimized to minimize their distance from the network centroid. As a result, the positions of the network graph nodes and the connections they define can be used to interpret the dimensions of the projected space and explain the positions of plotted points in the space. This network graph captures the dominant patterns of variation in ordered connections but does not represent the full high-dimensional structure of the network. The two dimensions correspond to the variance explained by the first two singular vectors used for visualization. As is standard in ENA/ONA, interpretation focuses on relative node positioning and connection patterns in this reduced space, while acknowledging that additional variation is distributed across higher dimensions not shown. Our model show a strong goodness of fit between the visualization and the original model -- correlations of 1 (Pearson) and 1 (Spearman) for the first dimension and co-registration correlations of 0.98 (Pearson) and 0.97 (Spearman) for the second. 

We construct two different ONA models to compare across reporting mechanisms (self- vs probe-caught) and task duration groups (shorter vs longer completion time, defined by a median split). Where possible, we also make use of ONA  difference networks, where node positions remain fixed while edge weights are color-coded to represent contrasts between conditions. Specifically, we report a difference model for task duration, but construct separate ONA networks for reporting differences, as a class imbalance between probe-caught reports (64\%) and  self-caught reports (36\%) would make a difference network difficult to interpret.  Additionally, we report the weights of the connections between nodes in each visualization. 

\section{Results}
\subsection{Descriptive Results}
\begin{figure}[ht]
\centering

\begin{minipage}[t]{0.48\textwidth}
\centering
\includegraphics[width=\linewidth]{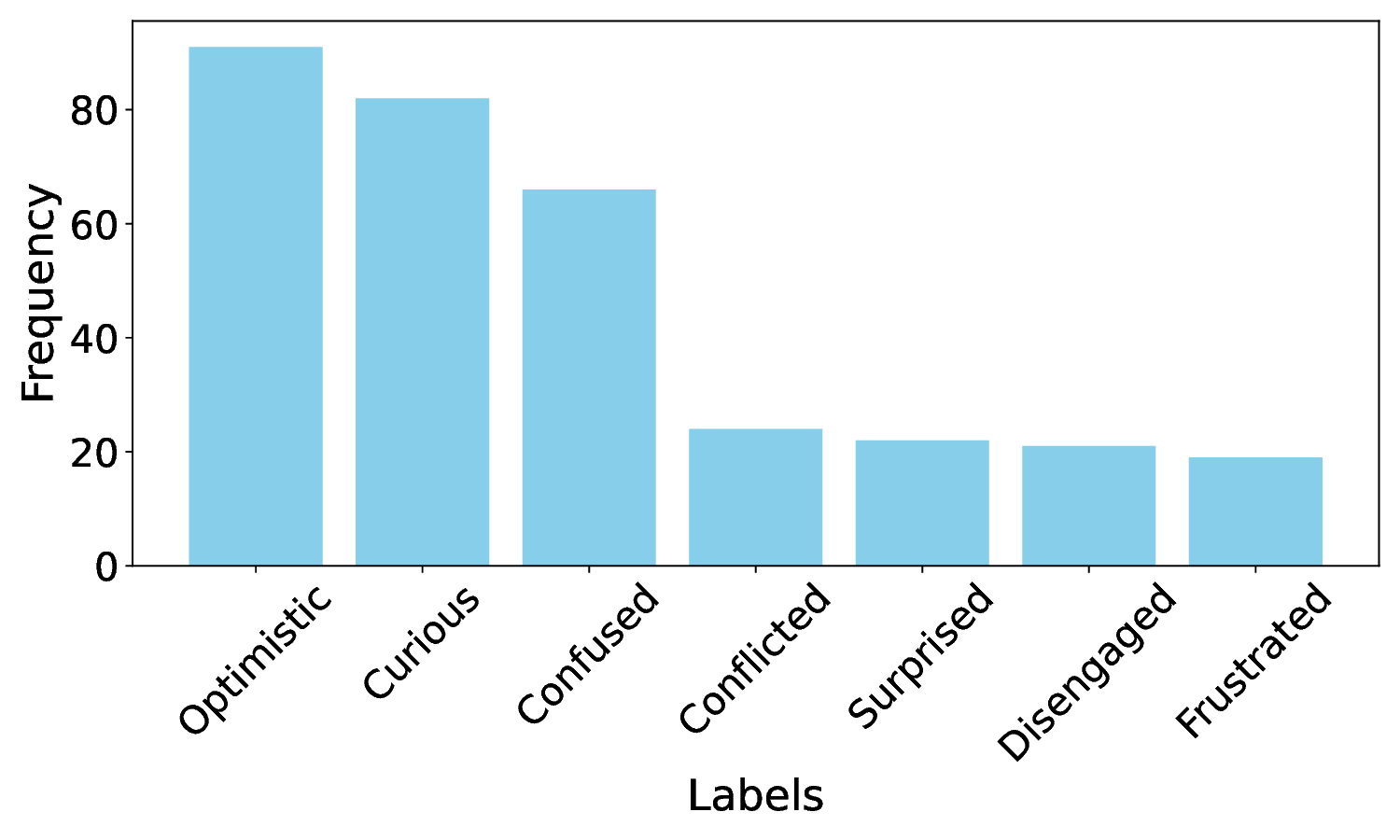}

\vspace{2mm}
{\small \raggedright \textbf{(a)} Overall frequency distribution of affective states across all reports.}
\end{minipage}
\hfill
\begin{minipage}[t]{0.48\textwidth}
\centering
\includegraphics[width=\linewidth]{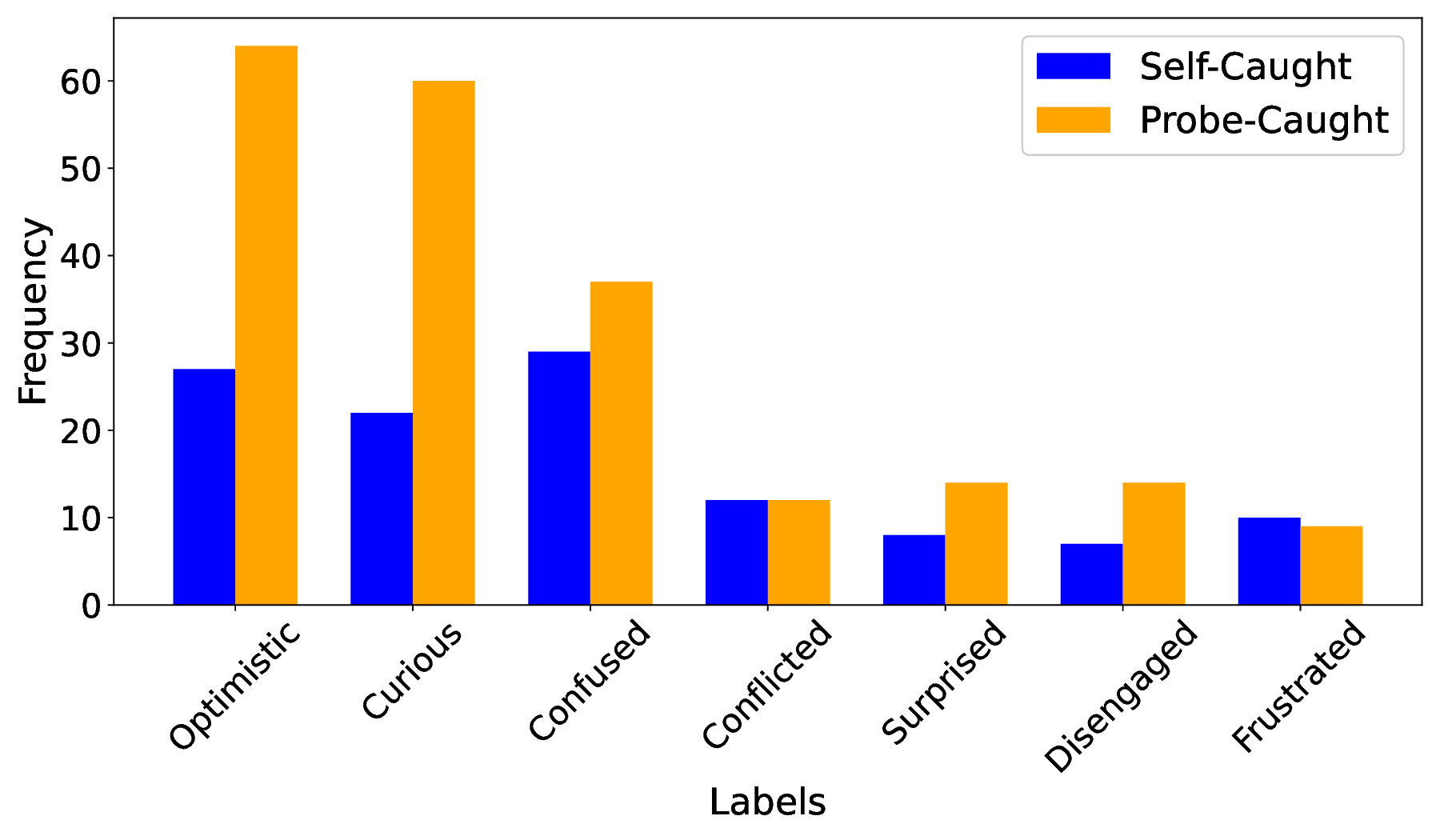}

\vspace{2mm}
{\small \raggedright \textbf{(b)} Frequency of affective states across reporting mechanisms.}
\end{minipage}

\caption{Frequency distributions of reported affective states.}
\label{fig:overall_freq}
\end{figure}

We first report the overall frequency distribution of affective states across all reports (Figure \ref{fig:overall_freq}(a)). Curious, optimistic, and confused were the most frequently reported states, followed by surprised, disengaged, frustrated, and conflicted. All 7 states appeared across groups, indicating coverage of the full label set in the dataset. 

Figure \ref{fig:overall_freq}(b) compares the frequency of affective states reported using self-caught and probe-caught methods. Both methods exhibited similar overall profiles, with curious, optimistic, and confused remaining the most prevalent states. However, self-caught reports were more evenly distributed across the set of affective states, while probe-caught reports showed greater concentration among a smaller subset of states, suggesting differences in which states are more salient under each reporting method.
\FloatBarrier
\subsection{ONA Analysis, Group Speed Differences}
\begin{figure}[ht]
\centering
\begin{minipage}[t]{0.49\textwidth}
\centering
\includegraphics[width=\textwidth]{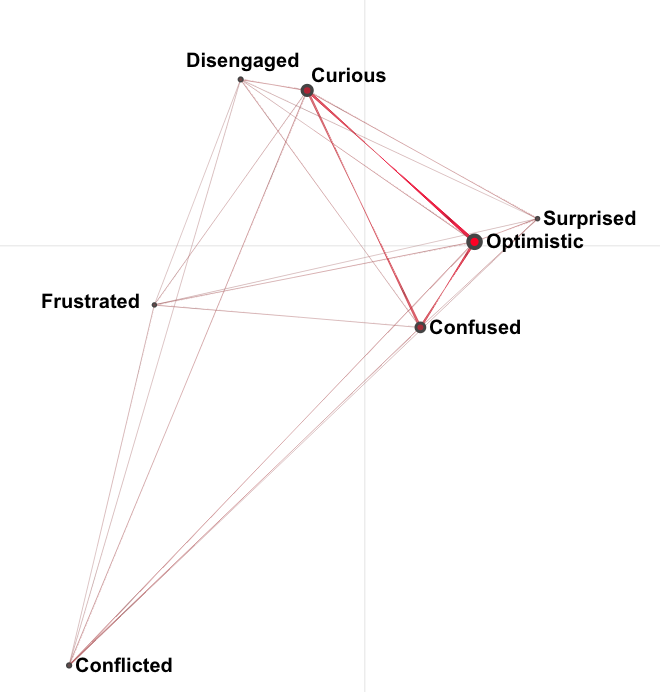}
\vspace{2mm}
{\fontsize{9}{11}\selectfont
\textbf{(a)} ONA network for affective states across all CPS groups. The first two dimensions explain 34.3\% and 21.3\% of the variance.}
\end{minipage}
\hfill
\begin{minipage}[t]{0.49\textwidth}
\centering
\includegraphics[width=\textwidth]{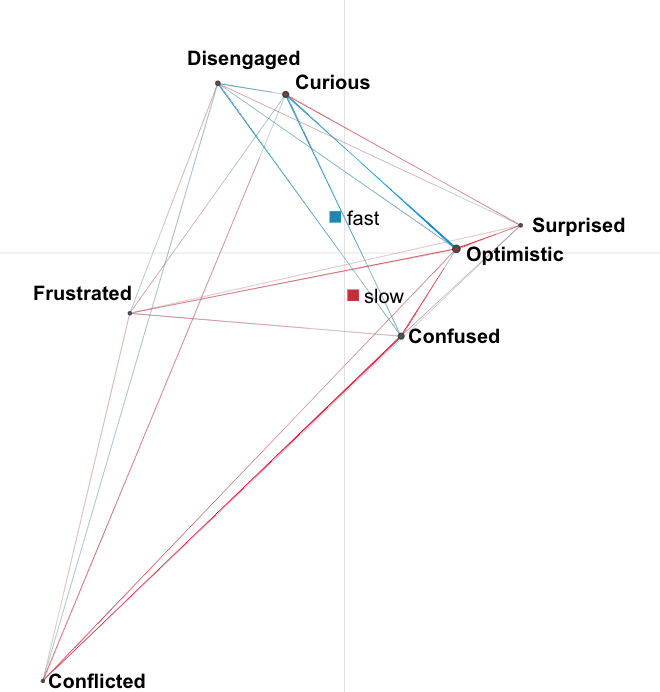}
\vspace{2mm}
{\fontsize{9}{11}\selectfont
\textbf{(b)} Difference network comparing slower (red) and faster (blue) groups. The first two dimensions explain 34.3\% and 21.3\% of the variance.}
\end{minipage}
\caption{Ordered network analysis (ONA) visualizations of affective state dynamics.}
\label{fig:ona_comparison}
\end{figure}

We use ordered network analysis (ONA) to compare the affective transitions of groups who finished the task more quickly to those who completed them more slowly. As the ONA with the full dataset shows (see Figure \ref{fig:ona_comparison}(a)), the primary connections are found among curious, optimistic, and confused. Additional connections involving conflicted, frustrated, surprised, and disengaged are present but comparatively weaker and more peripheral.
A closer inspection of the ONA model suggests that the horizontal dimension separates epistemic states associated with forward progress (e.g., curiosity, optimism, surprise) from those associated with difficulty and struggle (e.g., confusion, conflict, frustration). The vertical dimension is less readily interpretable as a single construct. A distinction between struggle (with frustrated, confused, and conflicted appearing lower in the graph) and disengagement (with disengagement in the highest location), appears to be emerging, but disengagement's proximity to curious makes this more challenging to interpret. 
% These interpretations are intended as descriptive heuristics derived from the projected structure and not validated constructs.
\begin{table}[H]
\centering
\caption{Ordered connection weights for affective states across all groups}
\label{tab:ona_aggregated}
\setlength{\tabcolsep}{10pt}
\renewcommand{\arraystretch}{1.1}
\vspace{1mm}
\begin{tabular}{lcccccccc}
\toprule
\textbf{From} & \textbf{CUR} & \textbf{OPT} & \textbf{CON} & \textbf{CFL} & \textbf{SUR} & \textbf{DIS} & \textbf{FRU} \\
\midrule
\textbf{CUR} & 0.24 & 0.23 & 0.18 & 0.07 & 0.07 & 0.08 & 0.04 \\
\textbf{OPT} & 0.31 & 0.39 & 0.24 & 0.07 & 0.06 & 0.07 & 0.06 \\
\textbf{CON} & 0.20 & 0.20 & 0.21 & 0.06 & 0.03 & 0.04 & 0.04 \\
\textbf{CFL} & 0.06 & 0.09 & 0.10 & 0.03 & 0.02 & 0.02 & 0.02 \\
\textbf{SUR} & 0.07 & 0.08 & 0.06 & 0.02 & 0.02 & 0.00 & 0.01 \\
\textbf{DIS} & 0.07 & 0.03 & 0.06 & 0.02 & 0.03 & 0.07 & 0.02 \\
\textbf{FRU} & 0.04 & 0.05 & 0.05 & 0.04 & 0.02 & 0.01 & 0.04 \\
\bottomrule
\end{tabular}
\end{table}
A qualitative analysis of the task recordings show that states positioned toward the right side of the space (e.g., optimism, surprise, confusion) tend to occur in moments where expectations or the problem state changes (such as realizing that a certain block is heavier than the group perceived it to be), whereas states on the left side (e.g., disengagement, conflict, frustration) occur around a broader range of experiences, including apparent boredom, tension, and disagreement. 
\begin{table}[H]
\centering
\caption{Ordered connection weights from the difference network in Figure \ref{fig:ona_comparison}(b). Positive values indicate stronger connections in slower groups; negative values indicate stronger connections in faster groups.}
\label{tab:ona_fast_slow_diff}
\setlength{\tabcolsep}{10pt}
\renewcommand{\arraystretch}{1.1}
\vspace{1mm}
\begin{tabular}{lcccccccc}
\toprule
\textbf{From} & \textbf{CUR} & \textbf{OPT} & \textbf{CON} & \textbf{CFL} & \textbf{SUR} & \textbf{DIS} & \textbf{FRU} \\
\midrule
\textbf{CUR} & 0.07 & -0.11 & -0.10 & -0.01 & 0.07 & -0.02 & -0.03 \\
\textbf{OPT} & -0.15 & 0.07 & -0.03 & 0.05 & 0.10 & -0.06 & 0.06 \\
\textbf{CON} & -0.05 & 0.09 & -0.04 & 0.11 & 0.03 & -0.05 & 0.02 \\
\textbf{CFL} & 0.05 & 0.07 & 0.00 & 0.00 & 0.00 & 0.01 & 0.00 \\
\textbf{SUR} & 0.04 & 0.06 & 0.03 & -0.01 & 0.02 & -0.01 & 0.01 \\
\textbf{DIS} & -0.07 & 0.00 & -0.09 & -0.02 & 0.02 & 0.04 & 0.02 \\
\textbf{FRU} & 0.03 & 0.00 & 0.04 & 0.02 & 0.01 & -0.01 & -0.06 \\
\bottomrule
\end{tabular}
\end{table}
Figure \ref{fig:ona_comparison}(b) presents the same data, but this time shows it as a difference network contrasting faster and slower groups, as operationalized by a median split. Blue edges indicate ordered transitions that were stronger in faster groups (negative values), whereas red edges indicate transitions that were stronger in slower groups (positive values).  Line weights for this (see Table \ref{tab:ona_fast_slow_diff}) shows, faster groups experience much stronger affective transitions among curious and optimistic, occasionally transitioning from optimistic to disengaged (-0.05).  In contrast, slower groups showed stronger bidirectional transitions among confused, conflicted, and frustrated. Particularly, these groups were much more likely to transition from confused to conflicted (0.11) than faster groups, and less likely to transition from confused to disengaged (-0.05). For reference, Table \ref{tab:ona_aggregated} shows the line weights with the full dataset.
% NOTE THAT TABLE THREE ONLY SHOWS DIFFERENCES
\FloatBarrier
\subsection{ONA Analysis, Self vs. Probe Initiated }

\begin{figure}[ht]
\centering

\begin{minipage}[t]{0.48\textwidth}
\centering
\includegraphics[width=\textwidth]{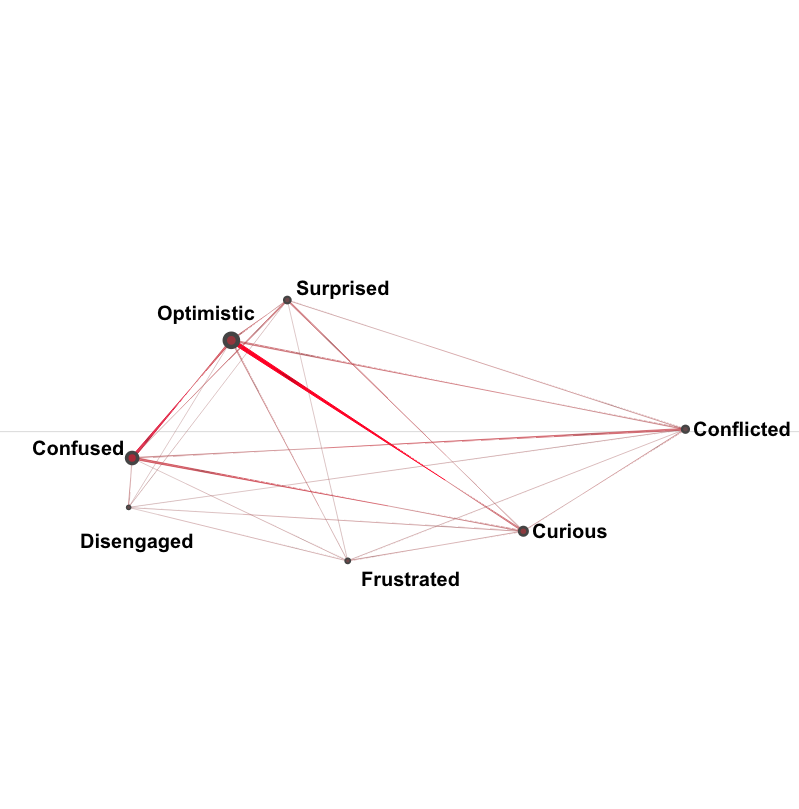}

\vspace{2mm}
{\fontsize{9}{11}\selectfont
\textbf{(a)} ONA model derived from self-caught. The X and Y axes describe 29.2\% and 16.5\% of the variance, respectively.}
\end{minipage}
\hfill
\begin{minipage}[t]{0.48\textwidth}
\centering
\includegraphics[width=\textwidth]{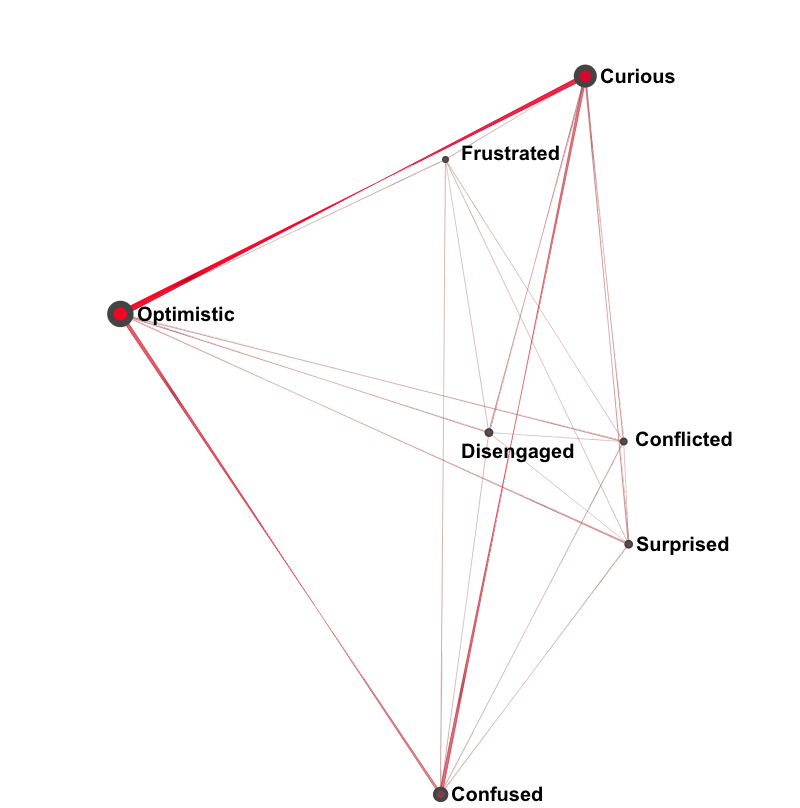}

\vspace{2mm}
{\fontsize{9}{11}\selectfont
\textbf{(b)} ONA model derived from probe-caught. The X and Y axes describe 31.1\% and 24.7\% of the variance, respectively.}
\end{minipage}

\caption{Ordered affective networks across self- and probe-caught reporting mechanisms.}
\label{fig:ordered_reporting_mechanisms}

\end{figure}
We next use an ONA analysis to examine differences in the type of self-reporting structure used by looking at the entirety of the dataset.  Because of a large class imbalance between self-initiated and probe-initiated self reporting, a difference model comparing the two would have been difficult to interpret. Instead we construct separate ONA models for the self-intiated transitions and the probe-initiated transitions, as shown in Figures \ref{fig:ordered_reporting_mechanisms}(a) and \ref{fig:ordered_reporting_mechanisms}(b) . The associated line weights are shown in {Table \ref{tab:ona_edges_combined}.

As these comparisons show, the strongest ordered connections continue to be found  among Curious, Optimistic, and Confused, which show strong bidirectional transitions and self-loops in both datasets. However, the model shapes of the two datasets diverge beyond that. 

In self-caught reports, the model appears rotated, but otherwise preserves the distinction between conflicted and the other affective states seen in the full data. However, the relationship between curious and confused appears to have inverted. Whereas curious was quite close to disengaged in the full data, that confusion now appears closest to it. 

In contrast, the shape of the ONA model constructed solely from probe-caught reports looks quite different. In this data, optimistic is the affective state most separated from the group (appearing to the far left of the axis), while confused and curious define the major distinctions of the Y axis. 
\begin{table}[ht]
\centering
\caption{Ordered connection weights reported as self-caught / probe-caught, with the stronger connection weights in \textbf{bold}.}
\label{tab:ona_edges_combined}
\setlength{\tabcolsep}{7pt}
\renewcommand{\arraystretch}{1.1}
\vspace{1mm}
\begin{tabular}{lcccccccc}
\toprule
\textbf{From} & \textbf{CUR} & \textbf{OPT} & \textbf{CON} & \textbf{CFL} & \textbf{SUR} & \textbf{DIS} & \textbf{FRU} \\
\midrule
\textbf{CUR} & .13/\textbf{.31} & .12/\textbf{.30}  & .08/\textbf{.16} & .04/.04 &  .05/\textbf{.08} & .05/\textbf{.08} & \textbf{.04}/.00 \\
\textbf{OPT} & .32/\textbf{.36} & .15/\textbf{.37} & .16/\textbf{.22} & \textbf{.11}/.05 & \textbf{.10}/.05 & .01/\textbf{.07} & \textbf{.07}/.04 \\
\textbf{CON} & .15/\textbf{.19} & \textbf{.22}/.14 & \textbf{.19}/.12 & \textbf{.07}/.04 & .04/.04 & \textbf{.04}/.02 & .01/\textbf{.04} \\
\textbf{CFL} & \textbf{.06}/.05 & \textbf{.07}/.06 & \textbf{.14}/.04 & .02/.02 & \textbf{.06}/.01 & \textbf{.03}/.01 & .00/\textbf{.02} \\
\textbf{SUR} & \textbf{.10}/.07 & .06/\textbf{.08} & \textbf{.10}/.04 & \textbf{.03}/.01 & \textbf{.04}/.02 & \textbf{.02}/.00 & .00/\textbf{.02} \\
\textbf{DIS} & .02/\textbf{.07} & .01/\textbf{.05} & \textbf{.06}/.02 & \textbf{.01}/.00 & \textbf{.02}/.01 & .02/\textbf{.07} & .00/\textbf{.01} \\
\textbf{FRU} & \textbf{.04}/.03 & .02/\textbf{.04} & .03/.03 & .03/.03 & .01/\textbf{.02} & \textbf{.03}/.00 & \textbf{.06}/.00 \\
\bottomrule
\end{tabular}
\end{table}

The differences between these figures should be interpreted cautiously, since each necessarily omits transitions. It is interesting that the self-caught model more closely mirrors the full data set, given that the probe-caught reports were considerably more frequent.  Although it shows relatively stronger transitions involving conflicted, frustrated, and surprised, it also shows fewer self-loops. In contrast, the sheer-volume of probe-caught reports of curious, optimistic, and confused, all of which were relatively likely to persist (as seen in the self-loops), stretched the probe-caught network considerably. As a result, conflicted and disengaged, which were at opposite ends in the other models, now appear quite close, despite having weak line weights connecting them. 

\FloatBarrier
\section{Discussion}
This study joins a small but growing body of work that uses ordered network analysis (ONA) to examine students affective dynamics, this time within the context of collaborative problem solving (CPS). Specifically, we examine how the speed of different groups and the affect  reporting mechanisms shape these dynamics. In doing so, we seek to better contribute to how social learning activities like CPS might impact affect self-reporting.

Across ONA models, we observe a stable epistemic core which links curiosity, optimism and confusion. These states were consistently involved in the strongest ordered connections, including self-transitions, indicating that affective experience during collaboration is often characterized by short-range persistence rather than frequent switching. This aligns with literature on epistemic emotions, which frames confusion and curiosity as productive states that support sustained sense-making rather than rapid affective fluctuation. 

Task duration differences provide an informative view of how affective organization varies across collaborative contexts. Importantly, task duration was treated here as a descriptive property of collaboration rather than a proxy for group performance, as speed alone does not necessarily reflect the quality of collaborative reasoning. Groups with shorter task durations exhibit ordered structures centered on curiosity and optimism, with disengagement appearing as a frequent but fleeting experience. In contrast, groups with longer task durations exhibit stronger connections between confused and conflicted and between curious and conflicted. This contrast highlights a key and underexplored distinction between confusion and disengagement in CPS: slower groups were characterized by sustained epistemic struggle, whereas faster groups showed patterns consistent with reduced affective friction. As a whole, this evidence suggests that faster groups achieved smoother coordination that reduced conflict-driven affect. However, the faster groups' frequent transitions into disengagement might also suggest group dynamics where one or two members drive the task while others are more reserved. 

To compare the affective states that were harvested using different reporting mechanisms, we constructed two different ONA models. ONA models constructed from probe-caught reports exhibit stronger self-loops among the core epistemic states, indicating more sustained epistemic engagement over time. In contrast, ONA models constructed from self-caught reports showed more distributed transitions that involved affective states typified by higher arousal, such as frustration, surprise, and conflict. These patterns suggest that the students were reporting different kinds of experiences when they were pushed to report compared to when they volunteered. For example, stronger self-loops in probe-caught reports may reflect the continued sampling of an ongoing state, whereas self-caught reports are more likely to occur at moments of perceived affective change.
The distribution of reports across conditions was imbalanced (64\% probe-caught vs.\ 36\% self-caught), which may also influence the relative density of connections.

While exploratory, these patterns challenge definitions of successful CPS collaboration and highlight the need to study how confusion and disengagement uniquely impact team management and pacing. From an educational perspective, AI systems that assist in CPS should be designed to recognize the difference between productive struggle that warrants silence and struggle-free disengagement that requires intervention. In particular, affect-aware systems should monitor local trajectories of these states over time, rather than triggering responses based on isolated affective signals and summary statistics.

\section{Limitations and Future Work}
This study has several limitations. First, the dataset is relatively small with a proportionately large male population and includes a fixed set of predefined labels, which may limit the generalizability of the findings. As an initial application of ordered network analysis (ONA) to retrospective cued-recall (RCR) data, this work provides a baseline for how specific affective states organize over time in situated collaborative problem solving. Future research should replicate these analyses with larger, more diverse samples and incorporate a broader range of affective categories.Second, while RCR is a valuable tool for capturing subjective experience, it is susceptible to recall bias and temporal imprecision, which may affect the accuracy of reported ordering of affective states and, consequently, the modeled transitions. Because determining the “ground truth” for affective states remains a significant challenge in the field, future studies should consider triangulating self-reports with physiological or behavioral signals. Finally, this study does not include direct measures of learning outcomes or performance beyond task completion time, as all groups arrived at the correct solution. As such, we are unable to examine how observed affective structures relate to learning or solution quality. Future work should investigate how ONA-derived representations of affective organization relate to learning outcomes, collaboration quality, and the effectiveness of adaptive interventions in affect-aware educational technologies.

\section{Conclusion}
This paper examines the dynamics of affective states during situated collaborative problem-solving (CPS) through the lens of ordered network analaysis (ONA). Our results show that affective experience in CPS is structured around short-range persistence and transitions among a small set of epistemic states, with a stable core linking curiosity, optimism, and confusion across conditions.  We find that faster groups have more self-loops and transitions among curiosity and optimism, with disengagement as a peripheral state, whereas slower groups showed stronger coupling among confusion, conflict, and frustration. These findings suggest that collaborative speed reflects differences in affective and epistemic organization rather than serving as a simple indicator of performance quality, and highlight distinct roles for confusion and disengagement in shaping collaborative trajectories. Comparisons between reporting mechanisms show that probe-caught reports have stronger self-loops and transitions among the stable core states, while self-caught is more distributed and interconnected with peripheral states. Through this work, we contribute both methodological and empirical knowledge on how social learning activities such as situated CPS impact the dynamics of affective states. We show that collaboration is tied to specific patterns of epistemic persistence and conflict, which has implications for how adaptive systems monitor group progress.
\begin{credits}
\subsubsection{\ackname}
This material is based in part upon work supported by Other Transaction award
HR00112490377 from the U.S. Defense Advanced Research Projects Agency
(DARPA) Friction for Accountability in Conversational Transactions (FACT)
program, and by the National Science Foundation (NSF) under a subcontract
to Colorado State University on award DRL 2454151 (Institute for Student-AI
Teaming). Approved for public release, distribution unlimited. Views expressed
herein do not reflect the policy or position of, the Department of Defense, the
National Science Foundation, or the U.S. Government. All errors are the responsibility of the authors. We thank Mariah Bradford and Dr. Nikhil Krishnaswamy for their assistance with data collection and to the anonymous reviewers whose
feedback helped improve the final copy of this paper.
\subsubsection{\discintname}
The authors have no competing interests to declare for this article.
\end{credits}

\begin{thebibliography}{8}
\bibitem{hesse_cps}
Hesse, F., Care, E., Buder, J., Sassenberg, K., Griffin, P.: A framework for teachable
collaborative problem solving skills. In: Griffin, P., Care, E. (eds.) Assessment and
Teaching of 21st Century Skills. EAIA, pp. 37–56. Springer, Dordrecht (2015).
https://doi.org/10.1007/978-94-017-9395-7\_2

\bibitem{dmello_dynamics}
D’Mello, S., Graesser, A.: Dynamics of affective states during complex learning. Learn. Instr. 22(145–157), 9 (2012)

\bibitem{ismail_emotionally_adaptive_2023}
Ismail, D., Hastings, P.: Emotionally adaptive intelligent tutoring system to reduce foreign language anxiety. In: Wang, N., Rebolledo-Mendez, G., Dimitrova, V., Matsuda, N., Santos, O.C. (eds.) Artificial Intelligence in Education (AIED 2023). CCIS, vol. 1831. Springer, Cham (2023). https://doi.org/10.1007/978-3-031-36336-8\_55
% Ismail, D., Hastings, P. (2023). Emotionally Adaptive Intelligent Tutoring System to Reduce Foreign Language Anxiety. In: Wang, N., Rebolledo-Mendez, G., Dimitrova, V., Matsuda, N., Santos, O.C. (eds) Artificial Intelligence in Education. Posters and Late Breaking Results, Workshops and Tutorials, Industry and Innovation Tracks, Practitioners, Doctoral Consortium and Blue Sky. AIED 2023. Communications in Computer and Information Science, vol 1831. Springer, Cham. https://doi.org/10.1007/978-3-031-36336-8\_55

\bibitem{yu_affect_prediction_2024}
Yu, H. et al.: Affect Behavior Prediction: Using Transformers and Timing Information to Make Early Predictions of Student Exercise Outcome. In: Olney, A.M., Chounta, IA., Liu, Z., Santos, O.C., Bittencourt, I.I. (eds) Artificial Intelligence in Education. AIED 2024. Lecture Notes in Computer Science(), vol 14830. Springer, Cham. https://doi.org/10.1007/978-3-031-64299-9\_14

% \bibitem{gupta_affect_tools_2021}
% Gupta, A. et al. (2021). Affective Teacher Tools: Affective Class Report Card and Dashboard. In: Roll, I., McNamara, D., Sosnovsky, S., Luckin, R., Dimitrova, V. (eds) Artificial Intelligence in Education. AIED 2021. Lecture Notes in Computer Science(), vol 12748. Springer, Cham. https://doi.org/10.1007/978-3-030-78292-4\_15

\bibitem{pekrun_emotions_measurement_2011}
Pekrun, R., Goetz, T., Frenzel, A. C., Barchfeld, P., \& Perry, R. P. (2011). Measuring emotions in students’ learning and performance: The Achievement Emotions Questionnaire (AEQ). Contemporary Educational Psychology

\bibitem{baker_bromp_2020}
% Baker, R. S. et al. (2020): BROMP quantitative field observations: A review. Learning science: Theory, research, and practice, 127-156.
Baker, R. S., Ocumpaugh, J. L., \& Andres, J. M. A. L. (2020). BROMP quantitative field observations: A review. Learning science: Theory, research, and practice, 127-156.

\bibitem{baker_sensor_free_affect_2012}
Baker, R.S.J.d., Gowda, S.M., Wixon, M., Kalka, J., Wagner, A.Z., Salvi, A., Aleven, V., Kusbit, G.W., Ocumpaugh, J. \& Rossi, L. (2012). Towards Sensor-Free Affect Detection in Cognitive Tutor Algebra. In International Conference on Educational Data Mining (EDM) 2012.

\bibitem{hussain_autotutor_physiology_2011}
Hussain, M.S. et al.: Affect Detection from Multichannel Physiology during Learning Sessions with AutoTutor. In: Biswas, G., Bull, S., Kay, J., Mitrovic, A. (eds) Artificial Intelligence in Education. AIED 2011. Lecture Notes in Computer Science(), vol 6738. Springer, Berlin, Heidelberg.

\bibitem{dmello_language_emotions_2012}
D’Mello, S.K., Graesser, A.: Language and discourse are powerful signals of student emotions during tutoring. IEEE Transactions on Learning Technologies 5(4), 304–317 (2012). https://doi.org/10.1109/TLT.2012.10

\bibitem{zhang_llm_think_aloud_srl_2024}
Zhang, J., Borchers, C., Aleven, V., Baker, R.S.: Using large language models to detect self-regulated learning in think-aloud protocols. In: Paassen, B., Demmans Epp, C. (eds.) Proceedings of the 17th International Conference on Educational Data Mining (EDM 2024), pp. 157–168. International Educational Data Mining Society, Atlanta, GA, USA (2024). https://doi.org/10.5281/zenodo.12729790

\bibitem{bosch_video_affect_2016}
% Bosch, N. et al.: Using video to automatically detect learner affect in computer-enabled classrooms. ACM Transactions on Interactive Intelligent Systems 6(2), 1–26 (2016).
Bosch, N., D’Mello, S.K., Ocumpaugh, J., Baker, R.S., Shute, V.: Using video to automatically detect learner affect in computer-enabled classrooms. ACM Transactions on Interactive Intelligent Systems 6(2), 1–26 (2016).

\bibitem{akbulut_self_reports_bias_2025}
Akbulut, Y. (2025). Beyond self-reports: Addressing bias and improving data quality in educational research. Journal of Measurement and Evaluation in Education and Psychology, 16(2), 115-123.

\bibitem{underwood_social_masking_1997}
Underwood, M.K.: Peer social status and children’s understanding of the expression and control of positive and negative emotions. Merrill-Palmer Quarterly 43, 610–634 (1997).

\bibitem{russell_retrospective_cued_recall_2009}
Russell, D.M., Oren, M.: Retrospective cued recall: A method for accurately recalling previous user behaviors. In: Proceedings of the 42nd Hawaii International Conference on System Sciences (HICSS 2009), pp. 1–9 (2009).

% \bibitem{rosenberg1994emotioncoherence}
% Rosenberg , E. and Ekman , P. 1994 . Coherence between expressive and experiential systems in emotion . Cognition and Emotion , 8 ( 3 ) : 201 – 229 .

% \bibitem{dmello2009halflives}
% D’Mello, S., \& Graesser, A. (2011). The half-life of cognitive-affective states during complex learning. Cognition and Emotion, 25(7), 1299–1308. https://doi.org/10.1080/02699931.2011.613668

\bibitem{bentley_rcr_gaming_2005}
Bentley, T., Johnston, L., \& von Baggo, K. (2005, November). Evaluation using cued-recall debrief to elicit information about a user's affective experiences. In Proceedings of the 17th Australia Conference on Computer-Human Interaction: Citizens Online: Considerations for Today and the Future (pp. 1-10).

\bibitem{anindho_internal_states_2025}
Anindho, S., Venkatesha, V., Bradford, M., Cleary, A.M., Blanchard, N. (2025). An Exploration of Internal States in Collaborative Problem Solving. In: Kurosu, M., Hashizume, A. (eds) Human-Computer Interaction. HCII 2025. Lecture Notes in Computer Science, vol 15768. Springer, Cham. https://doi.org/10.1007/978-3-031-93845-0\_11

\bibitem{tan_ona_2023}
Tan, Y., Ruis, A., Marquart, C., Cai, Z., Knowles, M., Shaffer, D.: Ordered Network Analysis. In: Damşa, C., Barany, A. (eds.) Advances in Quantitative Ethnography. ICQE 2022. Communications in Computer \& Information Science, vol. 1785 (2023)

\bibitem{pekrun_control_value}
Pekrun, R.: The control-value theory of achievement emotions: Assumptions, corollaries, and implications for educational research and practice. Educational Psychology Review 18(4), 315–341 (2006). https://doi.org/10.1007/s10648-006-9029-9

\bibitem{karumbaiah_self_transition_2019}
Karumbaiah, S., Baker, Ryan S., Ocumpaugh, J.: The case of self-transitions in affective dynamics. In: Isotani, S., Millán, E., Ogan, A., Hastings, P., McLaren, B., Luckin, R. (eds.) AIED 2019. LNCS (LNAI), vol. 11625, pp. 172–181. Springer, Cham (2019). https://doi.org/10.1007/978-3-030-23204-7\_15

\bibitem{ocumpaugh_sdvet_2025}
Ocumpaugh, J., et al.: Refocusing the lens through which we view affect dynamics: the skills, difficulty, value, efficacy and time model. Presented at the Learning Analytics and Knowledge Conference, Dublin, Ireland (2025). 

\bibitem{dmello_half_life_2011}
D'Mello, S., \& Graesser, A. (2011). The half-life of cognitive-affective states during complex learning. Cognition \& Emotion, 25(7), 1299-1308

\bibitem{baker_better_frustrated_2010}
Baker, R.S.J.d., D’Mello, S.K., Rodrigo, M.M.T., Graesser, A.C.: Better to be frustrated than bored: The incidence, persistence, and impact of learners’ cognitive–affective states during interactions with three different computer-based learning environments. International Journal of Human-Computer Studies 68(4), 223–241 (2010). https://doi.org/10.1016/j.ijhcs.2009.12.003
\bibitem{lui_sequences_frustration_2013}
Liu, Z. et al.: Sequences of frustration and confusion, and learning. In Proceedings of the Educational data mining 2013, Memphis, TN, USA, 6–9 July 2013.

\bibitem{shaffer2016quantitative}
Shaffer, D. W. et al.: A tutorial on epistemic network analysis: Analyzing the structure of connections in cognitive, social, and interaction data. Journal of learning analytics, 3(3), 9-45.

\bibitem{karumbaiah_ena_affect}
Karumbaiah, S., \& Baker, R. S. (2021, January). Studying affect dynamics using epistemic networks. In International Conference on Quantitative Ethnography (pp. 362-374). Cham: Springer International Publishing.

\bibitem{khebour_weights_task_2024}
Khebour, I. et al. (2024). When Text and Speech are Not Enough: A Multimodal Dataset of Collaboration in a Situated Task. Journal of Open Humanities Data, 10: 7, pp. 1–7. DOI: https://doi.org/10.5334/johd.168

% \bibitem{manty_socioemotional_cl_2020}
% Mänty, K., Järvenoja, H., \& Törmänen, T. (2020). Socio-emotional interaction in collaborative learning: Combining individual emotional experiences and group-level emotion regulation. International Journal of Educational Research, 102, 101589.

\bibitem{huang_social_emotion_cl_2023}
Huang, X., \& Lajoie, S. P. (2023). Social emotional interaction in collaborative learning: Why it matters and how can we measure it?. Social Sciences \& Humanities Open, 7(1), 100447.

\bibitem{nguyen_ssr_physiological_2023}
Nguyen, A. et al. (2023). Examining socially shared regulation and shared physiological arousal events with multimodal learning analytics. British Journal of Educational Technology, 54(1), 293-312.

\bibitem{dindar_physiological_arousal_cps_2022}
% Dindar, M. et al.: Detecting shared physiological arousal events in collaborative problem solving. Contemporary Educational Psychology, 69, 102050.
Dindar, M., Järvelä, S., Nguyen, A., Haataja, E., \& Çini, A. (2022). Detecting shared physiological arousal events in collaborative problem solving. Contemporary Educational Psychology, 69, 102050.

\bibitem{ma_confusion_detection_cl_2024}
Ma, Y. et al.: Automatically detecting confusion and conflict during collaborative learning using linguistic, prosodic, and facial cues. arXiv preprint arXiv:2401.15201 (2024)

\bibitem{webena}
Epistemic Network Analysis (ENA). https://www.epistemicnetwork.org/. Accessed 18 Jan 2026

\bibitem{bowman_ena_math_2021}
% Bowman, D. et al.: The mathematical foundations of epistemic network analysis. In A. Ruis \& S. B. Lee (Eds.), Advances in Quantitative Ethnography: Second International Conference, ICQE 2020, Malibu, CA, USA, February 1–3, 2021, Proceedings (pp. 91–105). Springer.
Bowman, D., Swiecki, Z., Cai, Z., Wang, Y., Eagan, B., Linderoth, J., \& Williamson Shaffer, D. (2021). The mathematical foundations of epistemic network analysis. In A. Ruis \& S. B. Lee (Eds.), Advances in Quantitative Ethnography: Second International Conference, ICQE 2020, Malibu, CA, USA, February 1–3, 2021, Proceedings (pp. 91–105). Springer.

% \bibitem{dmello_confusion_dynamics_graesser_2012}
% D’Mello, S., Graesser, A.: Confusion and its dynamics during learning. Learn. Instr. 22(3), 153–168 (2012)

\bibitem{chi_icap_2014}
Chi, M. T., \& Wylie, R. (2014). The ICAP framework: Linking cognitive engagement to active learning outcomes. Educational psychologist, 49(4), 219-243.
\end{thebibliography}
%

\end{document}